\documentstyle[icassp91]{article}

\title{DIALOGOS: A ROBUST SYSTEM FOR HUMAN-MACHINE SPOKEN DIALOGUE
ON THE TELEPHONE}

\name{{\normalsize Dario Albesano, Paolo Baggia, Morena Danieli,
Roberto Gemello, Elisabetta Gerbino, and Claudio Rullent}}

\address{{\small CSELT - Centro Studi e Laboratori Telecomunicazioni} \\
{\small Via G. Reiss Romoli 274, I-10148 Torino (Italy)} \\
{\small {\tt \{albesano, baggia, danieli, gemello, gerbino,
rullent\}@cselt.stet.it}}}

\input{psfig}

\begin{document}

\maketitle

\begin{abstract}

This paper presents Dialogos, a real-time system for 
human-machine spoken dialogue on the telephone in 
task-oriented domains. The system has been tested in a 
large trial with inexperienced users and it has proved 
robust enough to allow spontaneous interactions both to 
users which get good recognition performance and to the 
ones which get lower scores. The robust behavior of the 
system has been achieved by combining the use of 
specific language models during the recognition phase of 
analysis, the tolerance toward spontaneous speech 
phenomena, the activity of a robust parser, and the use of 
pragmatic-based dialogue knowledge. This integration of 
the different modules allows to deal with partial or total 
breakdowns of the different levels of analysis. We report 
the field trial data of the system and the evaluation results 
of the overall system and of the submodules.

\end{abstract}

\section{INTRODUCTION}

During the past few years the recognition of spontaneous 
speech in telephone dialogues has greatly improved. 
Nevertheless the natural spoken dialogue between 
computers and inexperienced users still presents some 
problematic issues, such as the real-time managing of 
large vocabularies, the robustness toward different 
pronunciations of a given natural language, and the 
ability of handling miscommunications within 
cooperative human-machine dialogues. Before delivering 
telephone-based spoken language applications to the 
general public, we have to define effective 
methodologies for overcoming these problems.

We present a telephone spoken dialogue system, 
Dialogos, that has been designed and implemented on the 
basis of the principle of strict integration among the 
different levels of analysis of user's utterances. That 
means that all the system modules are able to deal with 
partial or total breakdowns of the other modules.

Dialogos is a real time system that understands spoken 
Italian in the domain of railway timetable inquiry. It 
works on the public telephone network and it does not 
require any training to be used by inexperienced users. Its 
dictionary contains 3,471 words, including 2,983 proper 
names of the Italian railway stations.

The system is composed of a set of modules: the 
acoustical front-end, the acousting processor, the 
linguistic processor, the dialogue manager and the text-
to-speech synthesizer, which is the ELOQUENSâ 
commercial system by CSELT. A telephone interface 
connects the acoustical front-end and the synthesizer to 
the public telephone network, while the dialogue 
manager is connected to the railway timetable database. 
The telephone interface and the synthesizer are housed 
on a PC 486 equipped with Dialogic D41E boards. The 
railway time-table is on a PC Pentium and the rest of the 
system is software only and runs on a DEC Alpha 2100.

\section{ACOUSTIC PROCESSING}

The telephone signal, which has a band of 300-3400 Hz, 
is sampled at a frequency of 8 KHz. The pre-processing 
technique consists of a MEL-based spectral analysis 
followed by a Discrete Cosine Transform yelding a 
vector of 12 Cepstral Coefficients each 10 ms. In 
addition, the value of the logarithm of the total energy is 
retained as it provides some information about 
distinguishing the voiced parts of the speech from the 
unvoiced ones. First and second order derivatives of the 
log energy and of the 12 cepstral coefficients are also 
calculated resulting in a frame made up of 39 parameters.

The acoustic modeling is based on a hybrid HMM-NN 
(Hidden Markov Model-Neural Network) model~\cite{Gemello} of 
the same class as that described in~\cite{Bourlard}. Each word is 
described in terms of a left-to-right automaton (with self 
loops), obtained by concatenating elementary acoustic 
units. The posterior probability $P(Q|X)$ of the automata 
states are estimated by a Multi-layer Perceptron (MLP) 
neural network. The training of the acoustic model 
simultaneously finds the best segmentation of words into 
phonemes and of phonemes into states and trains the 
network to discriminate between these states.

Recently, Fissore et alii~\cite{Fissore} introduced a new set of 
units, called Stationary-Transitional Units (STU), which 
have been adopted instead of phonemes. These units are 
made up of stationary parts of the context independent 
phonemes plus all the admissible transitions between 
them for a total of 391 units. This set of STU is language 
dependent but domain independent, and represents a 
partition of the sounds of the language, like phonemes, 
but with more acoustic detail. The used MLP has one 
input layer that looks at 7 frames and two hidden layers. 
The output layer, fully connected, contains one unit for 
each STU. The total number of weights is 195,000.

The telephone quality speech used to train the HMM-
NN has the following features: 
\begin{itemize}
\item  read speech, domain independent, 1,136 speakers, 
about 8,000 utterances;
\item  spontaneous speech, domain dependent, about 3,580 
utterances
\end{itemize}

The recognition algorithm is based on frame 
synchronous Viterbi decoding. The recognition algorithm 
can work either in isolated or in continuous recognition 
mode and can be applied to different sets of words 
(vocabularies) to meet  the requirements of the dialogue 
manager.

\section{LANGUAGE MODELING}

The language model (LM) is a class-based bigram one. 
There are 358 classes; 348 of them contain a single word, 
while the remaining 10 classes contain semantically 
important words, such as city names (2,983 words), 
station names (33 words), numbers (76 words), months, 
week days, and so on.

The bigram model was trained on a set of 30,000 
sentences, which was composed of two parts: written 
material (86\%), and sentences acquired during a past trial 
(14\%). Currently the bigrams are smoothed using a linear 
interpolation algorithm, because the training set was too 
poor for performing other kinds of smoothing~\cite{Ney}.

Recently the use of dialogue-dependent prediction 
LMs have been integrated into the Dialogos system, 
see~\cite{Popovici}. These models are trained on a dialogue-de\-pen\-dent 
partition of a corpus acquired from a dialogue system 
according to the dialogue point in which an utterance was 
given. Our work is related to the static predictions of~\cite{Andry}
and to the dialogue step- dependent models of~\cite{Eckert}. On a 
test-set of 2,040 utterances, the use of dialogue-
dependent predictions reduces the error rate of WA by 
8.6\% and of SU by 10.9\%.

\section{LINGUISTIC PROCESSING}

The linguistic processor starts from the best-decoded 
sequence; it performs a multi-step robust partial parsing 
and, at the end of the analysis, it constructs the deep 
semantic representation of the user utterance in the form 
of a case frame and sends it to the dialogue module. The 
parser is designed to achieve robust performance; it is an 
evolution of the parser described in~\cite{Baggia}; studied to allow a 
faster definition of the linguistic knowledge to be used in 
application domains in the field of information inquiry. 
Only the grammatical structures that can give a 
contribution to the discrimination between different 
domain concepts conveyed by a given lexical item need 
to be defined and used.

Parsing is performed in three steps: a step of local 
grammatical analysis and two steps of semantic analysis. 
The grammatical analysis assigns to each lexical item a 
set of non terminals, that is, the union of the paths that in 
each syntactic tree connects that lexical item to the root. 
Notice that these trees do not necessarily cover the whole 
utterance: they are only the larger grammatical structures 
that include the given word. In addition, the trees 
pertaining to a lexical item do not necessarily cover the 
same utterance segment. To achieve robustness, local 
grammatical analysis is performed iteratively, starting 
from each word of the utterance and generating all the 
local grammatical structures that cover the utterance 
segments starting with such a word and being as long as 
possible. 

The grammar used to perform local grammatical 
analysis is written using a context-free like formalism; it 
is a 'semantic grammar' in the sense that the non-terminal 
names have to be defined considering not only syntactic 
knowledge but also a certain amount of semantic 
knowledge useful for the subsequent steps of semantic 
analysis.

The first step of the semantic analysis is completely 
local; it collects a set of application concepts, each one 
characterized by a score that represents the degree of 
linguistic reliability. The second step solves conflicts 
amongst these concepts and selects a set of mutually 
compatible application concepts.

\section{DIALOGUE MANAGEMENT}

The dialogue module (DM) has been designed to cope 
with task-oriented spoken langauage applications: that is, 
the DM performs its communicative actions to achieve 
the goal of collecting the parameters for accessing the 
database. At each turn of interaction with the user, the 
DM interprets the user's utterance on the basis of the 
dialogue history and of the contextual knowledge, and it 
selects a dialogue act that allows to address the user with 
a contextually appropriate message.

At each step of the human-machine interaction, the 
contextual knowledge of the DM is expressed in terms of 
pragmatic-based expectations about what the user could 
probably say in her/his next utterance. The possible 
discrepancies between the expectations of the system and 
the actual user's behavior are interpreted as symptoms of 
a breakdown in some previous steps of the ongoing 
interaction~\cite{Danieli}. When that happens, 
the system is able to 
continue the user-initiated repair. Moreover, the DM 
itself is able to initiate the recovering from other 
subcomponent errors both in case of total non-
understanding and in case of partial inconsistencies. 

Details of the implementation of the dialogue module 
are given in~\cite{Gerbino}. Briefly, the dialogue strategy of the 
DM assumes that both the user and the system cooperates 
for achieving the goal of their linguistic interchange. In 
our application domain that means that the user's goal 
and the system's goal converge to the identification of 
the parameters needed to access the data base, i.e. the 
departure and arrival cities, the date and the time of the 
travel. The DM prompts the user to provide such 
parameters, in an ordered fashion. However, the DM is 
able to deal with parameters which are relevant to the 
task and which are spontaneously offered by the user.

The DM interacts with the speech recognizer and with 
the database server. The interaction with the recognizer is 
implemented by passing to it the expectations of the DM 
in the form of predictions of class of words and phrases. 
Moreover, on the basis of the occurrence of repetitive 
recognition failures the DM may require the acquisition 
of some crucial parameters to be done in isolated speech 
recognition modality. 

The interaction with the database is bi-directional: on 
one hand, the DM simply sends to the database the 
queries as soon as the parameters involved have been 
acquired; on the other hand, it makes use of application 
dependent information for tailoring the dialogue strategy 
according to the kind of information actually needed to 
access the data-base.

There is an increasing aweraness that spoken language 
systems may greatly benefit from a robust dialogue 
management~\cite{Allen}. In a previous work~\cite{Danieli2}, 
we have identified two metrics (the explicit and the implicit 
recovery) that may be used to evaluate the robustness of 
the system by measuring the DM's ability to recover 
from miscommunications. By experimenting a previous 
version of the system with semi-naive and naive users, 
we deemed that the DM increased by 17\% the contextual 
appropriateness of the system answers.

\section{FIELD TRIAL EVALUATIONS}

An extensive field trial was carried out with 493 
Italian subjects. Subjects were recruited from all over 
Italy; they were statistically distributed, with regards to 
their regional origin, as the Telecom Italia users are. 
Subjects selected were roughly half male and half female, 
in the age range from 18 to over 65, and with different 
levels of education. 

Each subject had to do three telephone calls: in each 
one she/he had to plan a trip from a given city to another 
one. In the first call the subjects followed a pre-defined 
scenario that specified the departure and the arrival 
cities, while in the third call they were free to choose 
both the departure and the arrival point; in each one of 
the three calls they were free to decide the date and the 
time of departure.

The collected corpus consists of 1,363 dialogues for a 
total of 13,123 utterances. All the calls were performed 
over the public telephone network but in three different 
environments: house (80.3\% calls), telephone box (9.9\% 
calls) and some very noisy environments such as streets, 
cars, stations, and  underground (9.7\% calls). Four 
different kinds of telephone were used: DTMF phones 
used both in the house and telephone box (76.3\% calls), 
dial phones (8.1\% calls), cordless (5.9\% calls), and 
mobile phones (9.7\% calls). The mobile phones were 
always used in a noisy environment.

All the speech material acquired, 18 hours of speech, 
was manually transcribed and evaluated (487 Mbytes of 
data).

The dialogues have been evaluated both from the 
point of view of the overall system and from the point of 
view of the recognition and linguistic processing 
modules. With regards to the system's overall 
performance we classify each dialogue of the corpus into 
one of the following classes:
\begin{itemize}
\item {\sc SUCCESS} (S): complete successful dialogues: all the 
user parameters (departure, arrival, date, and time) 
have been correctly acquired and those parameters 
were used to access the database.
\item {\sc SUCCESS} with {\sc CONSTRAINT RELAXATION} (SC): 
successful dialogue where one parameter (date or 
time) was not recognized and the database is 
accessed with a default value, tomorrow for date and 
the main train connections of day for time.
\item {\sc SYSTEM FAILURE} (SF): dialogues that failed due to 
various kind of system inadequacies.
\item {\sc USER FAILURE} (UF): dialogues that failed due to a 
non-cooperative user behavior.
\end{itemize}

\begin{figure}[htb]
\centering
\psfig{figure=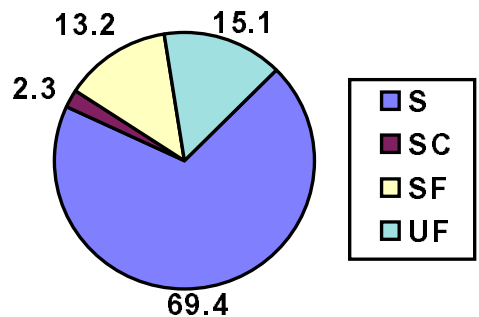,width=7.9cm}
\caption{\label{sts}Summary of Transaction Success}
\end{figure}

Figure~\ref{sts} shows the summary of transaction success: if 
we put together the S and SC dialogues we obtain the 
percentage of 71.7\% successful dialogues. If we exclude 
from the corpus the dialogues failed for user mistakes, 
we obtain the upper bound of the measure of transaction 
success, i.e. 84.4\%.

Analysing the three different scenarios, we can 
observe that users are able to adapt their speaking styles 
in order to be better understood by the system: they 
probably learn to speak after the tone. Both the users' 
and the system errors decrease from the first dialogue to 
the second, SF from 12.5\% to 10.3\% while UF from 
19.1\% to 14.3\%. In the third dialogue users continue to 
learn (their errors decrease to 12.0\%), but the system 
failures increase to 16.8\%, partially because the users 
asked connections for cities which were not present in 
the database.

We have also taken into account the different 
environments and telephone types used in the trial. It can 
be noticed that the DTMF telephone obtains the best 
results (S 85.5\%) while the dial phone obtains the worst 
results (S 77.1\%) and mobile phone, even if used in very 
noisy environment, obtains good results (S 80.0\%).

The average duration of the S dialogues is near to 2 
minutes. That time includes the readings of the retrieved 
railway information, which almost depends on the 
selected cities; 60\% of the S dialogues obtained the 
parameters to access the database in less than one minute.

We evaluated the 13,123 corpora sentences from the 
point of view of the recognition (word accuracy, WA) 
and understanding (sentence understanding, SU) 
performance; we obtain 61\% of WA and 76\% of SU. It is 
important to observe that 19\% of the utterances are 
affected by various kinds of spontaneous speech 
phenomena. In order of importance they are: shouts 
(4.7\% of sentences), restarts (5.1\% of sentences), 
extralinguistic phenomena (6.5\% of sentences), ill-
formed sentences (2.7\%) and out of dictionary words 
(5.7\% of sentences). 

By excluding these sentences the rate of WA and SU 
improves to 77.4\% and 83.6\% respectively.

\section{CONCLUSIONS}

The major advantage of Dialogos is its ability to allow a 
good level of efficiency for users that get good 
recognition performance, while the system relies on 
several recovery actions to allow most people with poor 
recognition performance to complete successfully their 
interactions. 

The experimental results show that most of the users 
were able to give and confirm all the required 
parameters, and that the system acquired those 
parameters with acceptable efficiency: 60\% of the users 
did that in less than one minute and 70\% in less than 
seven dialogue turns.

On the basis of the experimental data we can observe 
that the co-operative behavior by the user is essential: if 
we eliminate the non co-operative dialogues from the 
corpus, the rate of successful dialogues increases from 
71.7\% to 84.5\%. This datum suggests that in order to 
obtain realistic evaluations of spoken language systems 
performance, experimentation should migrate from the 
execution of realistic scenarios to the use of such systems 
by real users.

\begin{footnotesize}

\end{footnotesize}

\end{document}